\documentclass[11pt,a4paper]{article}
\usepackage[nohyperref]{acl2021}
\usepackage{times}
\usepackage{latexsym}

\usepackage{subfig}
\usepackage{graphicx}

\usepackage{microtype}

\aclfinalcopy 
\def\aclpaperid{1710} 

\usepackage{graphicx}
\usepackage{xspace}
\usepackage{paralist}
\usepackage{booktabs}
\usepackage{multirow}
\usepackage{amsmath}
\usepackage{amsfonts}
\usepackage{amssymb}
\usepackage{pifont}
\usepackage{mathrsfs}
\usepackage{algorithm,algpseudocode}
\usepackage{mdwlist}
\usepackage{enumitem}
\usepackage{color}
\usepackage[normalem]{ulem}
\usepackage{booktabs}
\usepackage{hyperref}

\newcommand{\cameraready}[1]{{ #1}}

\newcommand{\Method}{Shared Semantic Space\xspace}
\newcommand{\method}{shared semantic space\xspace}
\newcommand{\model}{Chimera\xspace}
\newcommand{\ModelFullname}{Text-Speech Shared Semantic Memory Network\xspace}
\newcommand{\modelfullname}{text-speech shared semantic memory network\xspace}
\newcommand{\methodprojection}{shared semantic projection\xspace}
\newcommand{\Methodprojection}{Shared semantic projection\xspace}
\newcommand{\MethodProjection}{Shared Semantic Projection\xspace}
\newcommand{\methodembedding}{memory query\xspace}
\newcommand{\methodembeddings}{memory queries\xspace}
\newcommand{\methodrepresentation}{semantic memory\xspace}
\newcommand{\methodrepresentations}{semantic memories\xspace}
\newcommand{\MethodRepresentations}{Semantic Memories\xspace}

\title{Learning \Method for Speech-to-Text Translation 
}

\author{Chi Han $^{1}$, Mingxuan Wang $^{1}$, Heng Ji $^{2}$, Lei Li $^{1}$ \\
    $^{1}$ ByteDance AI Lab, $^{2}$ University of Illinois at Urbana-Champaign \\
    {\tt \{hanchi.me, wangmingxuan.89, lileilab\}@bytedance.com} \\
    {\tt hengji@illinois.edu}
}

\date{}

\begin{document}
\maketitle

\begin{abstract}
    
Having numerous potential applications and great impact, end-to-end speech translation (ST) has long been treated as an independent task, failing to fully draw strength from the rapid advances of its sibling - text machine translation (MT). With text and audio inputs represented differently, the modality gap has rendered MT data and its end-to-end models incompatible with their ST counterparts.
In observation of this obstacle, we propose to bridge this representation gap with \model.
By projecting audio and text features to a common semantic representation, \model unifies MT and ST tasks and boosts the performance on ST benchmarks, MuST-C and Augmented Librispeech, to a new state-of-the-art. Specifically, \model obtains 27.1 BLEU on MuST-C EN-DE, improving the SOTA by a +1.9 BLEU margin.
Further experimental analyses demonstrate that the \method indeed conveys common knowledge between these two tasks and thus paves a new way for augmenting training resources across modalities.
Code, data, and resources are available at \url{https://github.com/Glaciohound/Chimera-ST}.    
\end{abstract}

\section{Introduction}
\label{sec:intro}

Speech-to-text translation (ST) takes speech input in a source language and outputs text utterance in a target language. It has many real-world applications, including automatic video captioning, simultaneous translation for international conferences, etc.
Traditional ST approaches cascade automatic  speech recognition (ASR) and machine translation (MT)~\cite{sperber2017neural,sperber2019self,zhang2019lattice,beck2019neural,cheng2019breaking}.
However, cascaded models often suffer from the issues of error propagation and translation latency.
As a result, there have been a series of recent attempts on end-to-end speech-to-text translation ~\citep{liu2019end, liu2018ustc, weiss2017sequence, berard2018end, duong2016attentional, jia2019leveraging,dong2021listen, wang2020curriculum}. The end-to-end approaches learn a single unified model, which is easier to deploy, has lower latency and could potentially reduce errors.

However, it remains a challenge for end-to-end ST to catch up with their cascaded counterparts in performance. We argue that the root cause is the gap between the two modalities, speech and text.
Although they both encode human languages, they are dissimilar in both coding attributes (pitch, volume, and intonation versus words, affixes, and punctuation)
and length (thousands of time frames versus tens of words). This issue is further coupled with the relatively smaller amount of parallel data for ST than for MT.


To tackle these challenges, we resort to making use of the additional available bilingual data for MT.
Our hypothesis is, to better leverage MT data,
an ideal model should be able to bridge the representations between speech and text.
Motivated by this intuition, we propose \model, a \modelfullname. It learns a \methodrepresentation by projecting features from both modalities into a \method.
This approach unifies ST and MT workflows and thus has the advantage of leveraging massive MT corpora as a side boost in training. It can also use speech-text pairs to align the \methodrepresentations from two modalities.

\cameraready{
This idea of a unified text-speech representation also finds its neural basis as suggested by recent evidence from functional neuroimaging
~\citep{vanatteveldt2004271, spitsyna2006converging, shankweiler2008reading}.
Specifically, ~\citet{vanatteveldt2004271, spitsyna2006converging} identifies certain regions in brain that the processing stream for speech sounds and visual texts converge at.
~\citet{shankweiler2008reading} further verifies that the size of such convergence sites correlates positively with the subjects' reading skills.
Coincidentally, at these convergence sites also found regions responsive to downstream activities such as lexical and semantical word recognition ~\citep{binder2003neural} and spontaneous generation of speech ~\citep{blank2002speech}. The evidence establishes the pivotal role of a modality-agnostic converged representation in language activities in brain.

This intuition lacks exploration in previous studies, with only a few exceptions ~\citep{indurthi2020data, liu2020bridging}, possibly due to the difficulties aforementioned and marginal improvements.}

Our results show that \model achieves new state-of-the-art results on all of 8 translation directions in the benchmark datasets MuST-C and Augmented LibriSpeech.
Specifically, \model obtains a 27.1 BLEU score on MuST-C EN-DE, which surpasses the best result ever reported by up to +1.9 BLEU. 
We also provide results under variations and ablations and validate our model design ideas by detailed analysis, as well as visualizing the semantic space \model has learned.

Our work makes the following contributions.
First, we propose \model, which is able to bridge the modality gap between speech and text. Second, we derive a novel bi-modal contrastive training task to learn an alignment between \methodrepresentations of two modalities. Finally, \model achieves a new state-of-the-art performance on the MuST-C benchmark and demonstrates its efficacy in learning modality-agnostic semantic representations.

\section{Related Work}
\label{sec:related}

\noindent\textbf{End-to-end ST}~
Since its first proof-of-concept work \citep{berard2016listen, duong2016attentional}, solving Speech Translation in an end-to-end manner has attracted extensive attention ~\citep{vila2018end, salesky2018towards, salesky2019fluent, di2019adapting, bahar2019comparative, di2019enhancing, inaguma2020espnet}.
Standard training techniques such as pretraining ~\citep{weiss2017sequence, berard2018end, bansal2019pre, stoian2020analyzing, wang2020bridging, pino2020self}, multi-task training ~\citep{vydana2021jointly, le2020dual, tang2021general}, meta-learning ~\citep{indurthi2020data}, and curriculum learning ~\citep{kano2018structured, wang2020curriculum} have been applied.
As ST data are expensive to collect, ~\citet{jia2019leveraging, pino2019harnessing, bahar2019using} augment synthesized data from ASR and MT corpora. Methods utilizing trained models, such as knowledge distillation ~\citep{liu2019end} and model adaptation ~\citep{di2020instance}, have also been shown to be effective.
Among these attempts, \cite{indurthi2020data, le2020dual, liu2020bridging}
are most related to ours, as they also attempt to train models on ASR or MT data. However, they both lack pivotal modules in model design to semantically bridge the gap between audio and text, and could thus suffer from modality mismatch in representations.

\noindent\textbf{Cascaded ST}~
The cascaded method is a more long-standing trend in ST
~\citep{sperber2017neural, jan2018iwslt}.
To alleviate its innate problem of error propagation, ~\citet{cheng2018towards, cheng2019breaking} introduce synthetic ASR-related errors and perturbations. On the other hand, some post-processing techniques such as re-segmentation ~\citep{matusov2006automatic}, punctuation restoration ~\citep{fugen2008system}, and disfluency detection ~\citep{fitzgerald2009reconstructing}
are proposed to fix flaws or errors that occurred during the translation.

\cameraready{
\noindent\textbf{Cross-Lingual Techniques}~
Techniques in multilingual tasks is also related to ours, as they aim at extracting common features out of sources from different representations (which, in this case, is language diversity) as well. However, multilingualism lacks key difficulties as observed in audio-text modality gap as discussed before.
~\cite{lu2018neural} and ~\cite{vazquez2019multilingual} are early attempts by building an LSTM-based attentional interlingua. ~\citet{yu2018multilingual, yang2019improving} uses a similar cosine-based loss for multilingual training. ~\citet{zhu2020language} is probably more similar in method to ours, but \model is more simple in terms of model and objectives, and the memories in \model are additionally designed to focus on specific semantic categories.
}

\section{Proposed Method: \ModelFullname}

\begin{figure*}[t!]
\centering
\includegraphics[width=0.9\textwidth]{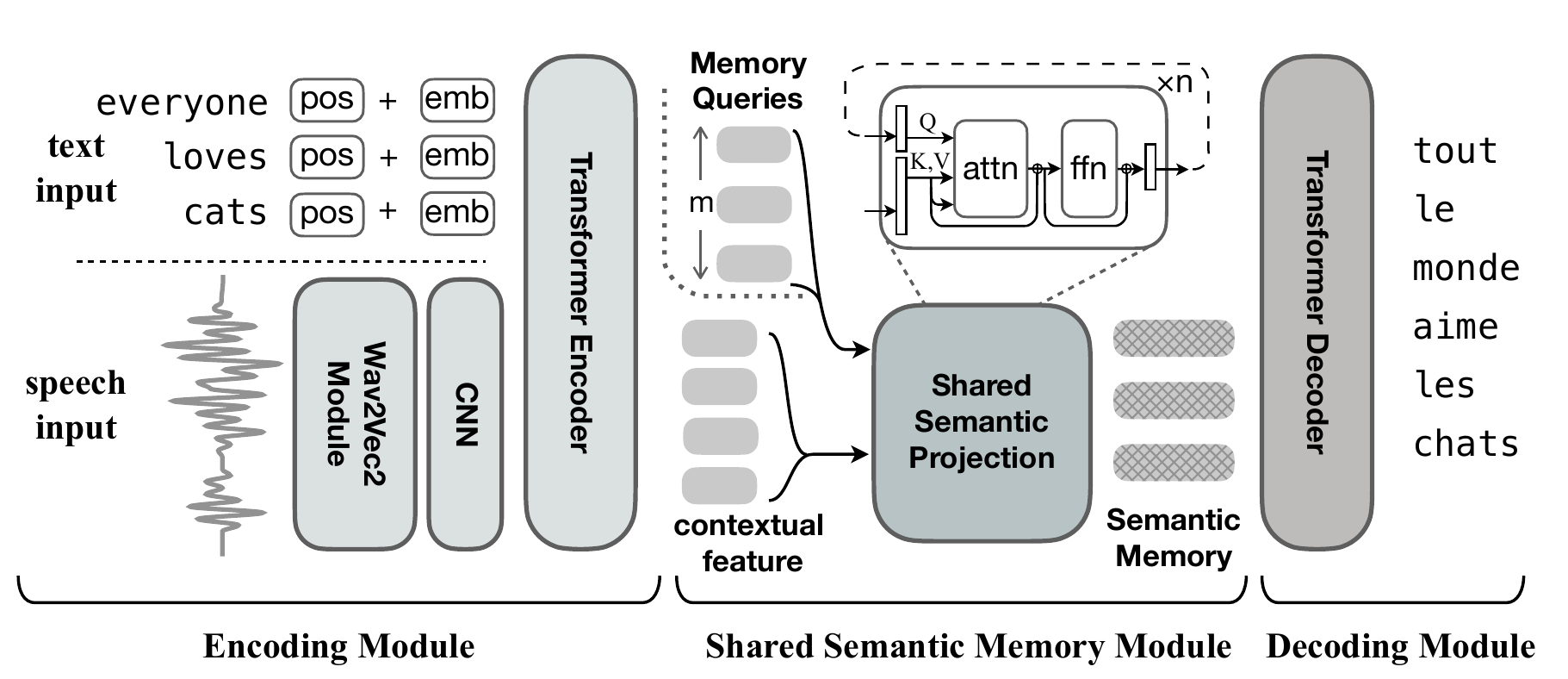}

\caption{An overview of the proposed \model. The Encoder Module contains Word embedding for text input, and Wav2vec2 sub-module for speech input. The \methodprojection Module uses its \methodembedding to produce \methodrepresentation with fixed-size representation from contextual features. The Decoder Module decodes translation from the \methodrepresentation.}

\label{fig:method}
\end{figure*}

\begin{figure}[t!]
\centering
\includegraphics[width=0.45\textwidth]{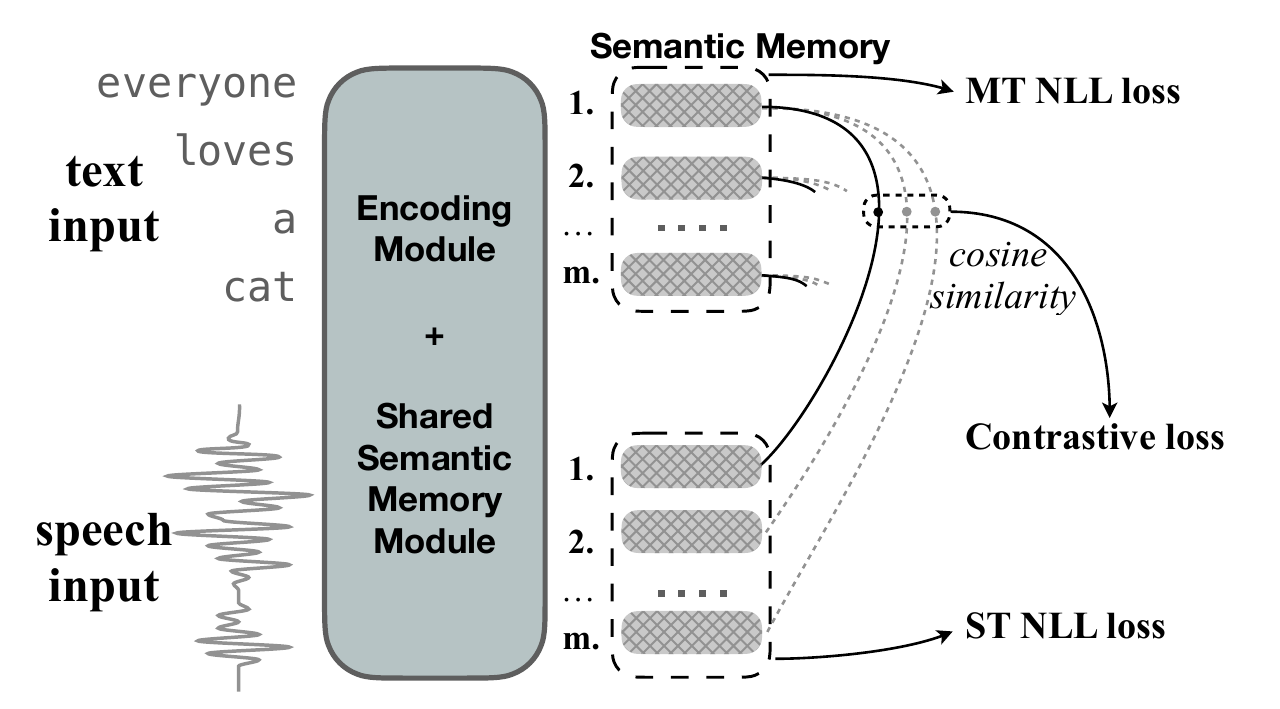}
\caption{Besides MT and ST translation loss, we adopt a bi-modal contrastive loss to help aligning the \methodrepresentations from text and speech. In short, among \methodrepresentation elements from both sides of paired speech and transcript, the contrastive loss maximizes the cosine similarity between the same \methodrepresentation element.}
\label{fig:contrastive}
\end{figure}
\label{sec:approach}


\subsection{Speech Translation Overview}
\label{subsec:overview}

An ST corpus usually consists of a set of triplet data \(\mathcal{S}=\{(\mathbf{x}_i, \mathbf{z}_i, \mathbf{y}_i)\}\). Here \(\mathbf{x}_i\) is the audio wave sequence, \(\mathbf{z}_i\) is the transcript sequence and \(\mathbf{y}_i\) is the translation sequence in the target language.
As a benefit of \methodprojection, \model is able to leverage large-scale MT training corpora \(\mathcal{T}=\{(\mathbf{u}_i, \mathbf{v}_i)\}\), where \(\mathbf{u}_i\) and \(\mathbf{v}_i\) are the source and target text sequences.

\subsection{\model Architecture}
\label{subsec:memory}

Figure \ref{fig:method} illustrates the structure of \model. It consists of three major components, an encoding module, a \methodprojection module, and a decoding module. 

\noindent\textbf{Encoding Module} 
Different from that of a conventional ST model, the encoding module of \model accepts either speech or text as input. For text input, we use word embeddings plus positional embeddings. For speech input, we use a pretrained Wav2Vec2 ~\citep{baevski2020wav2vec} to extract speech features. As the speech features can be very long, we apply an additional 1-dimensional strided CNN to reduce the length.
Both speech and text branches share a common subsequent Transformer encoder ~\citep{vaswani2017attention}. The final output of the encoding module is contextual features.

\noindent\textbf{\MethodProjection Module}
\Methodprojection plays a pivotal role in \model. The contextual features of speech and text may follow different distributions and of different lengths.
Ideally, the \methodprojection computes a constant number of semantic features as its output \methodrepresentations. 

This module take the contextual features out from the encoding module as input and then output \methodrepresentations of fixed length $m$. It consists of $n$ attentional layers.
It keeps a tuple of $m$ trainable input-dependent \methodembeddings to indicate the types of desired semantic information, which is used as the initial ``memories''. Uni-modal contextual features serves as attention ``keys'' and ``values'', while memories serves as attention ``queries``. Memories are iteratively fed to the $n$ \methodprojection layers, with each layer output used as input to next layer. The final output is used as the \methodrepresentation.

\begin{gather}
\mathbf{I}_0 = \mathbf{M}_0
\in \mathbb{R}^{m \times d}
\\
\mathbf{K}_i = \mathbf{V}_i = \mathbf{\hat{H}}
\in \mathbb{R}^{l\times d}
\\
\mathbf{I}_{i+1} = \mathbf{O}_i = \text{Attn}(\mathbf{I}_i, \mathbf{K}_i, \mathbf{V}_i)
\in \mathbb{R}^{m \times d}
\end{gather}
where $\mathbf{M}_0$, $\mathbf{\hat{H}}$, $\mathbf{I}_i$ and $\mathbf{O}_i$ denote the \methodembeddings, contextual features, the input, and the output of each layer, respectively. $l$ is the length of contextual features. $d$ is the shared vector dimension. The top-most output $\mathbf{O}_n$ is finally fed into the decoding module. 

\noindent\textbf{Decoding Module}
The decoding module contains a conventional Transformer decoder. The only difference is that the input is the fixed-size $\mathbf{O}_n$, which can possibly come from either speech or text.
\subsection{Training Objectives}
\label{subsec:obj}
The training objective of \model consists of three aspects, with their supervision signals coming from speech-to-text translation data $\{(\mathbf{x}_i, \mathbf{y}_i)\}$, text machine translation data $(\{\mathbf{u}_j, \mathbf{v}_j)\}$ and $(\{\mathbf{z}_i, \mathbf{y}_i)\}$ and the speech-transcript pairs $\{(\mathbf{x}_i, \mathbf{z}_i)\}$. 

\noindent\textbf{Speech-to-Text Translation Training}
The workflow of \model in Speech Translation is straightforward. The training objective is
negative log-likelihood on speech-to-text translation data $\{(\mathbf{x}_i, \mathbf{y}_i)\}$ from the data $S$ as the loss function.
\begin{equation} \label{eq:ST}
\mathcal{L}_\text{ST} = - \mathbb{E}_{\langle\mathbf{x},\cdot,\mathbf{y}\rangle \in \mathcal{S}} \log P (\mathbf{y} | \mathbf{x})
\end{equation}

\noindent\textbf{Text Machine Translation Training}
\model is also able to train on parallel sentences because of the unification of speech and text representations.
The parallel data are directly from MT dataset or transcript-translation pairs.
This enables the model to further acquire knowledge from the much larger extra MT corpus. 
Similar to ST training, the \methodprojection module projects the contextual text features to the \method, which are then taken by the decoding module to output a translation.
\begin{equation} \label{eq:MT}
\begin{aligned}
\mathcal{L}_\text{MT} = & - \mathbb{E}_{ \langle\cdot, \mathbf{z},\mathbf{y} \rangle \in \mathcal{S} } \log P (\mathbf{y} | \mathbf{z}) \\ 
& - \mathbb{E}_{\langle\mathbf{u},\mathbf{v} \rangle \in \mathcal{T}   } \log P (\mathbf{v} | \mathbf{u})
\end{aligned}
\end{equation}

\noindent\textbf{Bi-modal Contrastive Training}
The motivation of \model design is to bridge the speech and text representations.
We introduce dual-modal contrastive training to learn an alignment between representations from speech and text as illustrated in Figure \ref{fig:contrastive}.
First, \methodrepresentations from two inputs are computed. Then for each text \methodrepresentation, $\mathbf{M}_i^{text}$, we compute its cosine similarities with all speech \methodrepresentation  $\{\cos(\mathbf{M}_i^{text}, \mathbf{M}_j^{speech})\}$. They are then fed into a softmax function. The loss function maximizes the item from matched pairs $(\mathbf{M}_i^{text}, \mathbf{M}_i^{speech})$. Finally, the loss is summed across all text memory items and vise versa.


\begin{equation} \label{eq:contrastive}
\begin{split}
\mathcal{L}_\text{ctr} = &- \mathbb{E}_{\mathbf{x}, \mathbf{z}} \sum_i 
    \text{ln}  \frac{e^{\tau\text{cos} (\mathbf{M}_i^\text{text},\mathbf{M}_i^\text{speech})}}{\sum_j e^{\tau\text{cos} (\mathbf{M}_i^\text{text},\mathbf{M}_j^\text{speech})}} \\
    & -
    \mathbb{E}_{\mathbf{x}, \mathbf{z}} \sum_j 
    \text{ln}  \frac{e^{\tau\text{cos} (\mathbf{M}_j^\text{speech},\mathbf{M}_j^\text{text})}}{\sum_i e^{\tau\text{cos} (\mathbf{M}_j^\text{speech},\mathbf{M}_i^\text{text})}} 
\end{split}
\end{equation}

Intuitively, the contrastive loss forces the pair $(\mathbf{M}_i^{text}, \mathbf{M}_i^{speech})$ to project \methodrepresentations close to each other.
In the meantime, the softmax function trains the model to maintain diversity among \methodrepresentations.

The final loss is a weighted sum of each loss:
\begin{equation} \label{eq:loss}
\mathcal{L} =
\lambda_\text{ST} \mathcal{L}_\text{ST} +
\lambda_\text{MT} \mathcal{L}_\text{MT} +
\lambda_\text{ctr} \mathcal{L}_\text{ctr}
\end{equation}

\section{Experiments}
\label{sec:exps}

We conduct experiments on the benchmark MuST-C and, as a validation of model design, carry out ablation studies and visualize the representations \model has learned.

\begin{table*}[t!]
\centering
\small
\linespread{1}

\setlength{\tabcolsep}{1.5mm}{
\begin{tabular}{lccc|cccccccc}
\toprule
\multirow{2}{*}{\textbf{Model}} & \multicolumn{3}{c}{\textbf{External Data}} 
& \multicolumn{8}{c}{\textbf{MuST-C EN-X}}  \\
& Speech & ASR & MT 
& EN-DE & EN-FR & EN-RU & EN-ES & EN-IT & EN-RO & EN-PT & EN-NL\\




\midrule
FairSeq ST $^\dagger$
& $\times$ & $\times$ & $\times$ & 22.7 & 32.9 & 15.3 & 27.2 & 22.7 & 21.9 & 28.1 & 27.3 \\
Espnet ST $^\ddagger$
& $\times$ & $\times$ & $\times$ & 22.9 & 32.8 & 15.8 & 28.0 & 23.8 & 21.9 & 28.0 & 27.4 \\
AFS $^\star$
& $\times$ & $\times$ & $\times$ & 22.4 & 31.6 & 14.7 & 26.9 & 23.0 & 21.0 & 26.3 & 24.9 \\
Dual-Decoder $^\diamondsuit$
& $\times$ & $\times$ & $\times$ & 23.6 & 33.5 & 15.2 & 28.1 & 24.2 & 22.9 & \textbf{30.0} & 27.6 \\
STATST $^\sharp$
& $\times$ & $\times$ & $\times$ & 23.1 & - & - & - & - & - & - & - \\
MAML $^\flat$
& $\times$ & $\times$ & \checkmark & 22.1 & 34.1 & - & - & - & - & - & - \\
Self-Training $^\circ$
& \checkmark & \checkmark & $\times$ & 25.2 & 34.5 & - & - & - & - & - & - \\

W2V2-Transformer $^\ast$
& \checkmark & $\times$ & $\times$ & 22.3 & 34.3 & 15.8 & 28.7 & 24.2 & 22.4 & 29.3 & 28.2 \\

\hline

\model Mem-16 
& \checkmark & $\times$ & \checkmark & 25.6 & 35.0 & 16.7 & 30.2 & 24.0 & 23.2 & 29.7 & 28.5 \\

\model
& \checkmark & $\times$ & \checkmark & \textbf{27.1}$^\mathsection$ & \textbf{35.6} & \textbf{17.4} & \textbf{30.6} & \textbf{25.0} & \textbf{24.0} & \textbf{30.2} & \textbf{29.2} \\


\bottomrule
\end{tabular}}

\caption{Main results on tst-COMMON subset on all 8 languages in MuST-C dataset. ``Speech'' denotes unlabeled audio data. $^\mathsection$: the EN-DE result uses a mixed WMT14+OpenSubtitles data for MT pre-training. Among the baselines, $^\dagger$ shows results from ~\citet{ott2019fairseq}, $^\ddagger$ from ~\citet{inaguma2020espnet}, $^\star$ from ~\citet{zhang2020adaptive}, $^\diamondsuit$ from ~\citet{le2020dual}, $^\sharp$ from ~\citet{liu2020bridging}, $^\flat$ from ~\citet{indurthi2020data}, and $^\circ$ from ~\citet{pino2019harnessing}. $^\ast$ shows results of a simple baseline model by combining a Wav2Vec2 module ~\citep{baevski2020wav2vec} and a Transformer.
}
\label{tab:main_results}
\end{table*}
\begin{table}[t!]
\centering
\setlength\tabcolsep{3pt}
\begin{tabular}{cccc|cc}
\toprule

\multirow{2}{*}{\textbf{Model}} &
\multicolumn{3}{c}{\textbf{External Data}} &  \multicolumn{2}{c}{\textbf{Aug-Libri}} \\
& Speech & ASR & MT & u-tok & c-detok \\
\midrule

W2V2-T $^\ast$ & \checkmark & $\times$ & $\times$ & 6.5 & 6.4 \\
TCEN $^\dagger$ & $\times$ & $\times$ & $\times$ & 17.1 & - \\
LSTM $^\ddagger$& $\times$ & \checkmark & \checkmark & 17.0 & - \\
AFS $^\circ$ & $\times$ & $\times$ & $\times$ & 18.6 & 17.2 \\
Multilingual $^\star$ & $\times$ & \checkmark & $\times$ & 17.6 & - \\
Transformer $^\bot$ & $\times$ & \checkmark &  $\times$ & 17.7 & - \\
Curiculum $^\bot$ & $\times$ & \checkmark &  $\times$ & 18.0 & - \\
COSTT $^\flat$ & $\times$ & $\times$ & \checkmark & 18.2 & - \\
LUT $^\diamondsuit$ & $\times$ & \checkmark & $\times$ & 18.3 & - \\
STAST $^\sharp$ & $\times$ & \checkmark & $\times$ & 18.7 & - \\
\midrule
\model & \checkmark & $\times$ & \checkmark & \textbf{20.2} & \textbf{19.4} \\

\bottomrule

\end{tabular}
\caption{\cameraready{Results on LibriSpeech English-French dataset. Among baselines, 
$^\ast$ is the same W2V2-Transformer baseline as in Table \ref{tab:main_results}. 
$^\dagger$ is from ~\citet{wang2020bridging}, $^\ddagger$ from ~\citet{bahar2019using}, $^\star$ from ~\citet{inaguma2019multilingual}, two baselines under $^\bot$ from ~\citet{wang2020curriculum}, $^\flat$ from ~\citet{dong2021consecutive}, $^\diamondsuit$ from ~\citet{dong2021listen}, $^\circ$ from ~\citet{zhang2020adaptive} and $^\sharp$ from ~\citet{liu2020bridging}. ``u-tok'' represents case-insensitive tokenized BLEU, while ``c-detok'' is case-sensitive detokenized BLEU.
}}

\label{tab:librispeech}
\end{table}
\begin{table}[t!]
\centering
\small
\linespread{1}

\setlength{\tabcolsep}{1.5mm}{
\begin{tabular}{lccc|c}
\toprule
\multirow{2}{*}{\textbf{Model}} & \multicolumn{3}{c}{\textbf{External Data}} 
& \textbf{MuST-C}  \\
& Speech & ASR & MT &  EN-DE \\
\midrule

W2V2-T + Dec PT
& \checkmark & $\times$ & WMT14 & 22.2 \\

W2V2-T + KD
& \checkmark & $\times$ & WMT14 & 24.6\\

\model
& \checkmark & $\times$ & WMT14 & \textbf{26.3}  \\

\bottomrule
\end{tabular}}

\caption{
\cameraready{
Baselines that utilize external MT data. ``Dec PT'' pre-trains decoder on MT corpus; ``KD'' adopts the knowledge distillation technique as ~\citet{liu2019end}
}}
\label{tab:mt_baselines}
\end{table}

\subsection{Dataset and Preprocessing}
\label{subsec:data}

\noindent\textbf{MuST-C} ~\citep{di2019must} is a multilingual speech translation corpus with triplet data sources: source audio, transcripts, and text translations. 
MuST-C contains translations from English (EN) to 8 languages: Dutch (NL), French (FR), German (DE), Italian (IT), Portuguese (PT), Romanian (RO), Russian (RU), and Spanish (ES). With each pair consisting of at least 385 hours of audio recordings, 
to the best of our knowledge, MuST-C is currently the largest speech translation dataset available for each language pair. It includes data from English TED talks with manual transcripts and translations at the sentence level. We use the dev and tst-COMMON sets as our development and test data, respectively.

\cameraready{
\noindent\textbf{Augmented LibriSpeech Dataset (En-Fr)} ~\citep{kocabiyikoglu2018augmenting} is composed of aligned e-books in French and their human reading in English.
It provides typical triplet data of English speech, transcript and French text.
Following the setting of ~\citep{liu2019end}, we utilize the 100h hours of clean train set as training data, and use the original 2 hours of dev set and and 4 hours of test set.
}

\noindent\textbf{Machine Translation Datasets} After bridging the modality gap, \model has the potential power to utilize Machine Translation resources. Therefore we incorporate data from WMT, OpenSubtitles ~\citep{lison2016opensubtitles2016} and OPUS100~\citep{zhang2020improving} translation tasks. Specifically, we use WMT 2014 ~\citep{bojar2014findings} \footnote{http://www.statmt.org/wmt14/translation-task.html} for EN-DE, EN-FR, EN-RU and EN-ES,  WMT 2016 ~\citep{bojar2016findings} \footnote{ https://www.statmt.org/wmt16/translation-task.html} for EN-RO, and OPUS100 \footnote{http://opus.nlpl.eu/opus-100.php} for EN-PT, EN-IT, and EN-NL, as pretraining corpora. We additionally evaluate OpenSubtitles as EN-DE MT data to test the impact of MT corpus selection. WMT 2014 dataset provides at least 4 million sentences of translation data in each language pair. WMT 2016 contains less, around 600k for EN-RO direction. OPUS100 has around 1M sentences for each sentence pair. OpenSubtitles provides 22M sentences for EN-DE.

\noindent\textbf{Pre-processing of Data and Evaluation}
For speech input, the 16-bit raw wave sequences are normalized by a factor of $2^{15}$ to the range of $[-1, 1)$.

For text input, on each translation pair, all texts (ST transcripts and translation, and MT source and target texts) are pre-processed in the same way. Texts are case-sensitive. Punctuation is kept, split from words, and normalized. Non-print punctuation is removed. The sentences are then tokenized with Moses tokenizer \footnote{https://github.com/moses-smt/mosesdecoder/blob/master/scripts/tokenizer/tokenizer.perl}. We filter out samples whose number of source or target tokens is over 250 and whose ratio of source and target text lengths is outside range $[ 2/3, 3/2 ]$. For sub-wording, we use a unigram sentencepiece\footnote{https://github.com/google/sentencepiece} model with a dictionary size of 10000. On each translation direction, The sentencepiece model is learned on all text data from both ST and MT corpora. The dictionary is shared across MT and ST and across source and target languages.

Unless otherwise stated, performance is evaluated with BLEU~\citep{papineni2002bleu} using sacreBLEU \footnote{https://github.com/mjpost/sacrebleu, with configuration of 13a tokenzier, case-sensitiveness and full punctuation}.
We average 7 consecutive checkpoints around the one of the best dev loss and adopt a beam size of 10. 
\begin{table}[t!]
\centering
\begin{tabular}{cc|cc}
\toprule

S.S. Projection & Decoder & EN-DE & EN-FR \\ \midrule
 - & - & 25.6 & 35.0 \\
 Fixed & - & 24.3 & 34.3 \\
 - & Fixed & 24.2 & 33.4 \\
 Fixed & Fixed & 23.8 & 33.1 \\
 \bottomrule

\end{tabular}
\caption{Performance of Mem-16 \model when freezing different modules in fine-tuning. S.S. Projection is abbreviation for \methodprojection. ``Fixed'' indicates that weights in this module are not updated during fine-tuning, and ``-'' means otherwise. The results demonstrate that freezing modules indeed hampers the model's ability to adapt, but the weights pretrained on MT are already highly informative for ST.}
\label{tab:freezing}

\end{table}

\subsection{Model Configuration}
\label{subsec:config}

For text input, we use 512-dimensional word embeddings plus sinusoidal positional embeddings.
For audio input, the Wav2Vec2 Module follows the base configuration in ~\citet{baevski2020wav2vec}. It uses parameters pretrained on LibriSpeech audio data only. The 1-dimensional CNN for speech features has 2 layers with stride size 2, kernel size 5, padding 2, and hidden dimension 1024.

The shared Transformer encoder consists of 6 layers. The \methodembeddings are 64 512-dimensional vectors. The parameters of \methodprojection resemble a 3-layer Transformer encoder. The Transformer decoder has 6 layers.
Each of these Transformer layers, except for those in the Wav2Vec2 module, has an embedding dimension of 512, a hidden dimension of 512, and 8 attention heads.

In both pretraining and fine-tuning stages, we use an Adam optimizer with $\beta_1=0.9, \beta_2=0.98$, and 4k warm-up updates. We apply an inverse square root schedule algorithm for the learning rate.
In MT pretraining, the learning rate is 5e-4, the maximum number of updates is 300k, with at most 33k input tokens per batch.
In ST pretraining, the learning rate is 1e-4, the maximum number of updates is 150k, with at most 16M source audio frames per batch.
The loss weights $\lambda_\text{ST}$, $\lambda_\text{MT}$ and $\lambda_\text{ctr}$ are all set to 1.

We also show results on a base version of \model, for which the \methodembeddings are only 16 512-dimensional vectors (codenamed ``Mem-16''). Because of the training efficiency and simplicity, all ablation studies and visualizations adopted the Mem-16 configuration if not stated otherwise. 

Both \model and \model Mem-16 contain around 165M parameters. The whole training process for one trial on 8 Nvidia Tesla-V100 GPUs generally takes 20 \textendash 40 hours according to the translation direction.

\begin{table}[t!]
\centering
\begin{tabular}{cc|cc}
\toprule

MT & Contrastive & EN-DE & EN-FR \\ \midrule
 \checkmark & \checkmark & 25.6 & 35.0 \\
 \checkmark & $\times$ & 25.0 & 34.6 \\
 $\times$ & \checkmark & 24.7 & 34.6 \\
 $\times$ & $\times$ & 25.1 & 34.6 \\
 \bottomrule

\end{tabular}
\caption{BLEU scores of Mem-16 \model on MuST-C tst-COMMON set without one or both of auxiliary tasks. ``$\times$'' means this task is not used during fine-tuning, and ``\checkmark'' means othersize. ``Contrastive'' is the bi-modal contrastive task. The removal of one or both of tasks greatly harms the model's performance on both language pairs.}
\label{tab:multitask}

\end{table}

\subsection{Benchmark Experiments}
\label{subsec:benchmark}

\noindent\textbf{Training}
We train \model in a pretrain - fine-tune manner. 
In the first stage, we pretrain \model on MT datasets so as to leverage additional sources of training data, as well as provide a better initialization point.
In the fine-tuning stage, we adopt multi-task training as described in Section\ref{subsec:obj}. In addition to the conventional ST task, \model is also fine-tuned on MT and bi-modal contrastive task to align inputs from speech and text.


\noindent\textbf{Baselines} 
We include as baselines the speech transformer model from~\cite{ott2019fairseq}, Espnet result from~\cite{inaguma2020espnet}, adaptive feature selection method from~\cite{zhang2020adaptive}, dual-decoder Transformer from~\cite{le2020dual} and Modality-Agnostic Meta-Learning from~\cite{indurthi2020data} in Table \ref{tab:main_results}.
We also provide a series of baseline results of a simple combination of Wav2Vec2 ~\citep{baevski2020wav2vec} and Transformer.
It could be viewed \model without external MT pre-training, with still competitive but not SOTA results. 

To verify the effectiveness of our training technique, we also compare with other baselines able to leverage external MT corpus in Table \ref{tab:mt_baselines}.

\noindent\textbf{Results}
The experiment results are shown in Table \ref{tab:main_results} and \ref{tab:librispeech}. Our \model achieves state-of-the-art performance on all language pairs, even though we do not utilize Google Translate results on Augmented Librispeech as most baselines. EN-DE results of \model uses WMT14+OpenSubtitles for MT pre-trainng, while a detailed ablation study on the effect of MT data can be found in Section \ref{subsec:ablation_mtdata}. Note that the improvement on EN-PT is not so significant as EN-DE and EN-FR. We attribute this to the data discrepancy between OPUS100 and MuST-C. A large number of sentences in OPUS100 are from movie subtitles, which are more informal, contain repeated sentences, and cover different topics from those in MuST-C public speeches.

In Table \ref{tab:mt_baselines}, under the same data condition, \model outperforms other techniques such as decoder pre-training and knowledge distillation ~\citep{liu2019end}.


\begin{figure}[t!]
\centering

\subfloat[a][Chimera Mem-16 on WMT14]{
\includegraphics[width=0.45\textwidth]{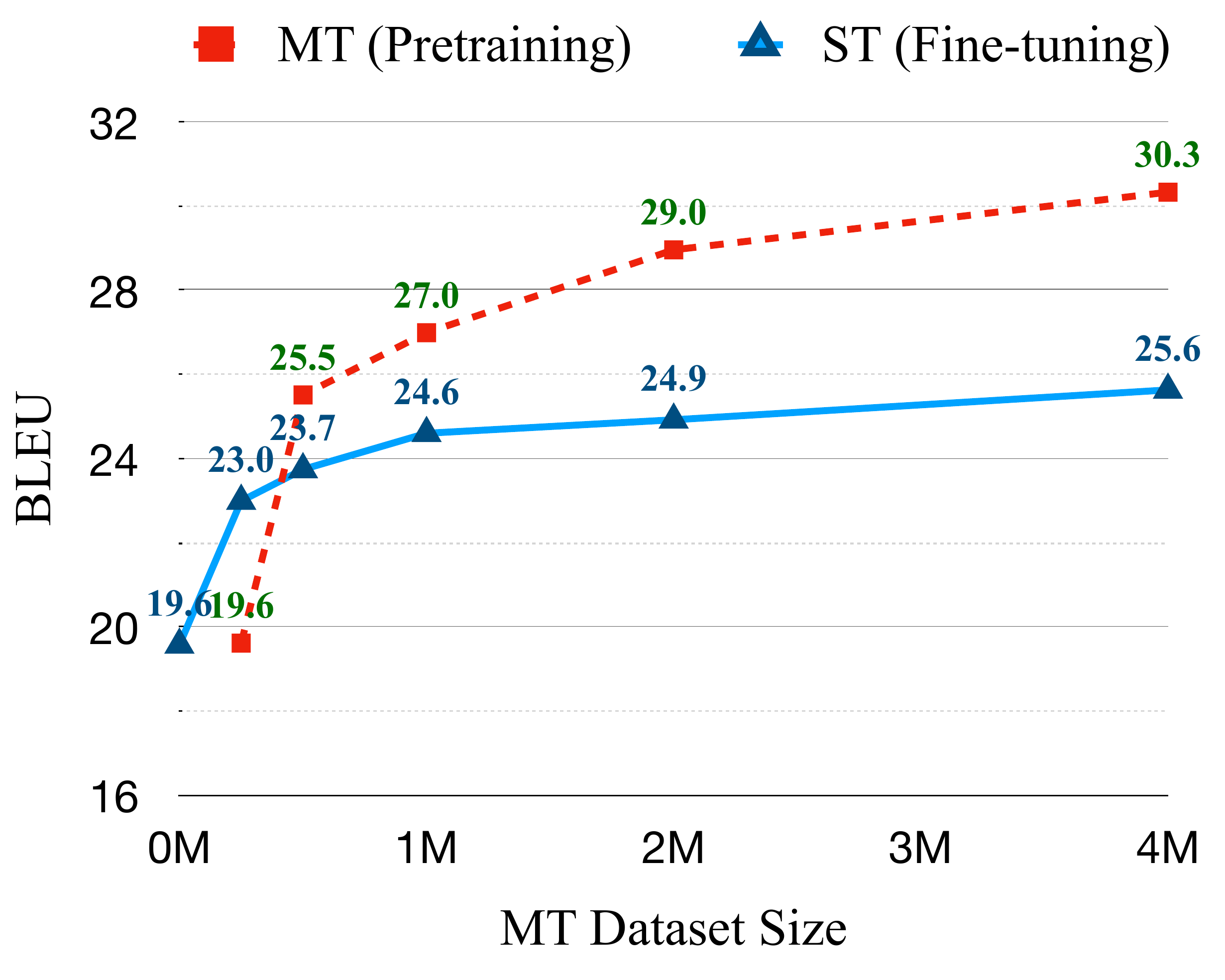}
}

\subfloat[c][Chimera w/ and w/o OpenSubtitles]{
\begin{tabular}{cc|cc}
\toprule

Corpora & Size & MT & ST \\ \midrule
WMT14 & 4M & 32.1 & 26.3 \\
WMT14 + OpenSubtitles & 26M & 32.9 & 27.1 \\
\bottomrule

\end{tabular}
}

\caption{
\cameraready{
Curve of MuST-C EN-DE tst-COMMON BLEU scores on \model against the amount of MT data used during pretraining. (a) shows \model Mem-16's performance on WMT14. Blue triangles are the speech translation BLEU scores, and green squares are transcript-translation BLEU scores after MT pretraining. (b) shows how \model ($M=64$) behaves with or without OpenSubtitles data.
}
}
\label{fig:mtdata}
\end{figure}


\subsection{Ablation Studies and Visualizations}
\label{subsec:ablation}

\begin{figure}[t!]
\centering
\includegraphics[width=0.5\textwidth]{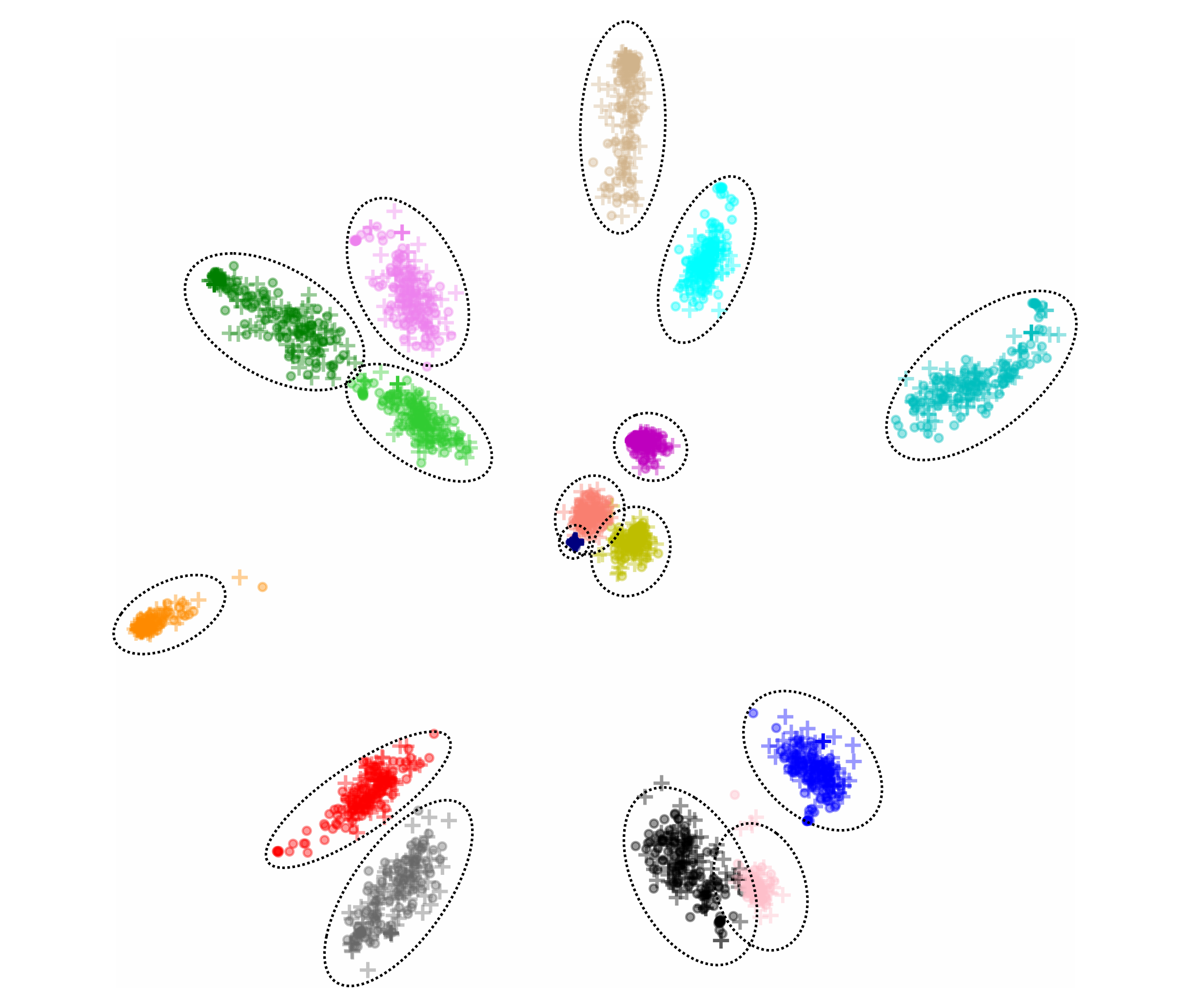}

\caption{2-dimensional PCA projection of the \methodrepresentations in Mem16 \model across different samples. Each colored cluster (circled out) represents a \methodrepresentation element,. A ``$\cdot$'' corresponds to a speech \methodrepresentation , and a ``+'' marks a text one.}

\label{fig:16memories}
\end{figure}

\begin{figure*}[t!]
\centering
\includegraphics[width=0.7\textwidth]{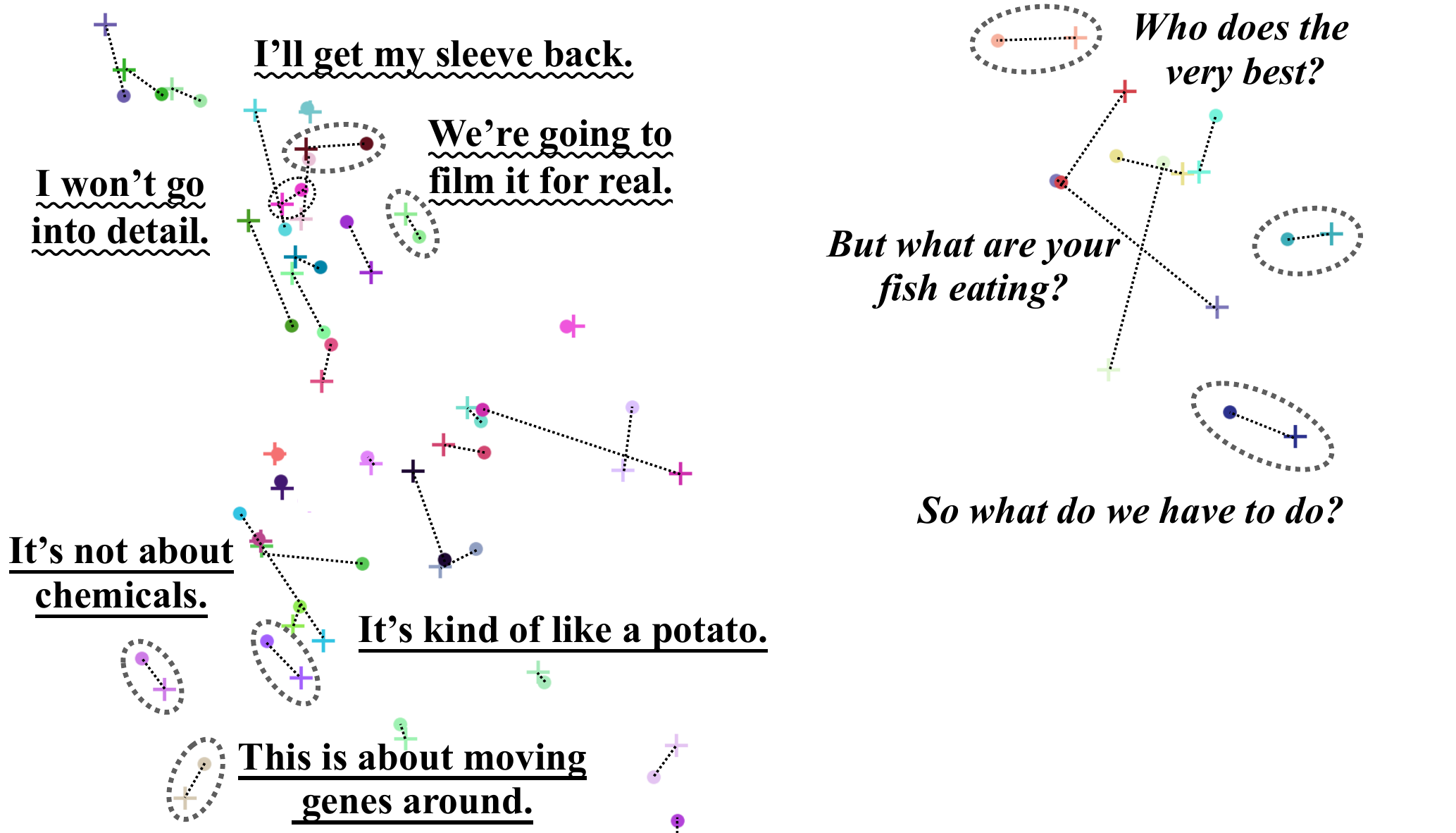}

\caption{A visualization of one particular \methodrepresentation in Mem-16 \model no different samples and modalities. ``$\cdot$'' marks speech representations, and ``+'' marks text representations. Marks of the same color come from the same speech-transcript pair and are linked with dashed lines. Some of speech-transcript pairs are circled together and annotated with their transcripts. Three fonts distinguish three groups of transcripts of similar patterns.}

\label{fig:visualization}
\end{figure*}

\paragraph{Knowledge Shared across Tasks}

One potential benefit in our design is that the \method can hold common knowledge shared across ST and MT tasks. To validate this motivating idea, we analyze the model's behavior while manipulating its modules. If certain weights pretrained during the MT task also contain meaningful information for ST, fixing them should not greatly harm the model's performance.

Specifically, after MT pretraining, we fix certain modules and do not update their weights during fine-tuning.
We choose to fix the weights in the \methodprojection module, the decoding module, or both of them.

Table \ref{tab:freezing} shows the results. After freezing modules, the results on both EN-DE and EN-FR drop slightly. This demonstrates that freezing weights indeed hampers the model's ability to adapt from MT to ST dataset. But the decreased scores are still comparable to many of the best results in Table \ref{tab:main_results}. This validates the effectiveness of \method, and indicates that the weights pretrained on MT are already informative enough for \model to still generalize sufficiently well on ST task.

\paragraph{Multi-task Training}

One advantage of bridging the modality gap is that the model can fully benefit from training on auxiliary tasks with more data, such as those mentioned in Section \ref{subsec:memory}.
To evaluate their impacts, we conduct another ablation study on EN-DE and EN-FR.. Either or both of the auxiliary tasks are not used during fine-tuning.

The results of this ablation are presented in Table \ref{tab:multitask}. Here we can see a significant decrease (with, for example, p=0.020 in one-tailed Student's t-test comparing row 1 and 2) in BLEU scores when either of the auxiliary tasks is abandoned. Although the bi-model contrastive loss is not directly related to the ST task, a poor alignment between \methodrepresentations of speech and text hinders \model from leveraging the shared knowledge across two tasks.
When the MT task is removed, the drop in BLEU scores is also huge.
This could be explained that during fine-tuning, the auxiliary MT task is necessary for keeping shared knowledge from being forgotten.

It is interesting to observe that abandoning both tasks produces results similar to abandoning either one of the tasks. This suggests that the two auxiliary tasks only have effects when combined with each other: only when both using MT task to maintain the pretrained parameters from forgetting, as well as using the bi-modal contrastive task to align between speech and text representations, can \model benefit from shared knowledge in MT pretraining.

\begin{figure*}[t!]
\centering
\includegraphics[width=\textwidth]{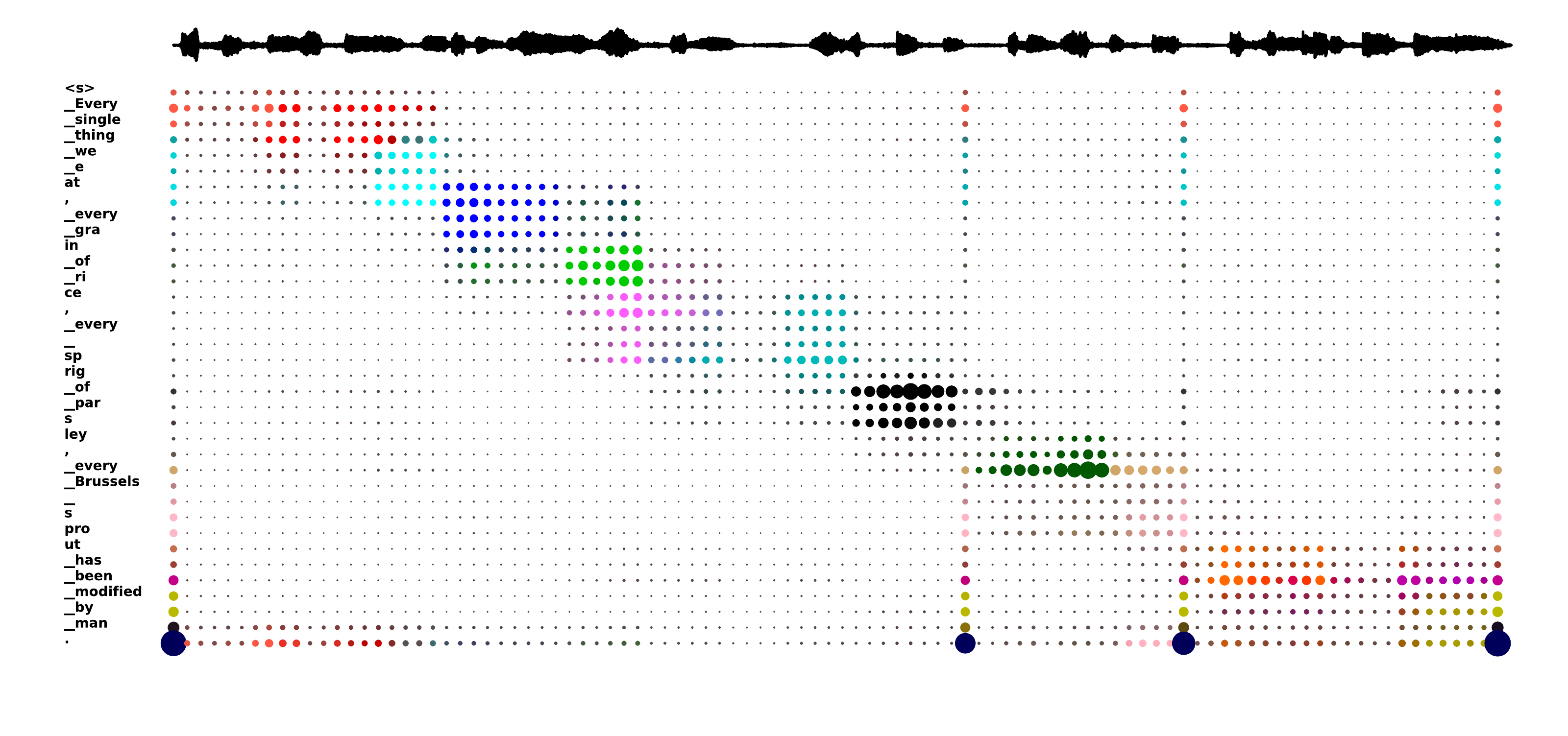}
\vspace{-2em}
\caption{
\cameraready{
Visualization of the final-layer attention of $M=16$ memories on inputs, and their alignment between each other. A pair of audio and its transcript is fed to \model Mem-16. The area of each dot is linear to dot product between two attention vectors. The color is a linear interpolation of $M$ indicative colors as in Fig \ref{fig:16memories}, with mixing weights linear to the Hadamard product of attention vectors. (Best viewed in color)
}
}

\label{fig:attention}
\end{figure*}
\paragraph{Additional Machine Translation Data}

\label{subsec:ablation_mtdata}

We attribute the gain in the performance of \model mainly to pre-training on MT data. One evidence is the performance gain when using the larger OpenSubtitles as MT corpus for EN-DE in \ref{tab:main_results}.
We also vary the amount of MT data available during pretraining on EN-DE direction. The results are plotted in Figure~\ref{fig:mtdata}. As the size of additional MT dataset increases, the MuST-C BLEU score improves significantly. This confirms the importance of massive high-quality MT data for pretraining \model. The results also help explain the relatively inferior scores on EN-PT in Table \ref{tab:main_results} which uses the OPUS100 dataset in pretraining.

\paragraph{Visualization of \MethodRepresentations}


The \methodprojection is designed to only extract semantic categories of information necessary for decoding, regardless of the input modality. In this way, it can bridge the different representations of speech and text during computation and facilitate knowledge sharing between MT and ST.

To validate this motivation, we visualize them with Principal Component Analysis (PCA) in Figure \ref{fig:16memories}. Up to 100 speech-transcript paired samples are randomly chosen from the valid set. 
We record vector values of 16 \methodrepresentations from Chimera Mem-16 when inputs are speeches or transcripts, and apply 2-dimensional PCA. The 16 \methodrepresentations are distinguished by 16 colors.
Every ``$\cdot$'' corresponds to a \methodrepresentation from speech, and each ``+'' is a \methodrepresentation from the text. It is clear that the \methodrepresentations are highly clustered, everyone of which learns a particular region. Speech and text representations are also projected close within the same region, proving the model's ability to ignore representation differences and bridge the modality gap.

To take a closer view of the structure of each \methodrepresentation subspace, we randomly choose one \methodrepresentation and apply PCA to its corresponding cluster. The results are visualized in Figure \ref{fig:visualization}.
These samples come from 50 speech-transcript pairs.
Each pair of speech (``$\cdot$'') and transcript (``+'') share the same color and are linked through dashed lines.

Two interesting properties could be observed.
First, paired speech and transcript inputs are again close to each other, even though they are coming from different modalities.
Second, the visualized representations are organized according to their semantic or syntactic patterns. We recognize several clusters in the figure, and annotate their transcripts with different fonts. The three annotations at the upper-right corner (Italic font) are all questions; those at the upper-left corner (wavy underlined font) all follow a simple future tense; at the bottom-left corner of the figure (underlined font) is another cluster of sentences of copular verbs. This proves that the \method that \model has learned is well-structured, and thus validates our model design.

\paragraph{Visualization of Inter-Modal Attention Alignment}

\cameraready{
"Attention" is the internal mechanism of Transformer based modules. In the design of \model, attention is used for extracting $M$ key semantic categories of features from input. To investigate whether these extracted features is indeed semantic, we further visualized the similarity between attention on paired audio and text in Fig \ref{fig:attention}. Here the colors, which distinguish different memories attending to inputs, is clustered on sequence and distributed close to the diagonal, demonstrating an alignment between matching tokens in two modalities. Here we also observe four beaming columns, where the full stop mark in text aligns with pauses in audio. This is an indication of semantic rather than positional essence of the memories.
}

\section{Conclusions and Future Work}
\label{sec:conclusion}

In this paper, we propose \model, a model capable of learning a \modelfullname for bridging the gap between speech and text representations. Being able to leverage a large amount of external Machine Translation data, \model achieves new state-of-the-art performance on the MuST-C dataset on all 8 languages. Additional experiment results also demonstrate its ability to learn a well-structured \method as well as effectively share learned knowledge across MT and ST, and validate our design of auxiliary tasks.

In the future, we will focus on deriving a better task to tightly align speech and text representations. Also, the workflows of MT and ST are only partially shared in \model, which still requires the model to adapt to ST when switching to the fine-tuning stage. So it remains a challenge to better couple their computation graphs in future designs.

\section*{Acknowledgments}
We would like to thank Rong Ye, Qianqian Dong, and Chengqi Zhao for many constructive suggestions and discussion about the ideas.

\bibliographystyle{acl_natbib}
\bibliography{paper}

\end{document}